\documentclass[10pt,twocolumn,letterpaper]{article}

\usepackage{cvpr}
\usepackage{times}
\usepackage{epsfig}
\usepackage{epstopdf}
\usepackage{graphicx}
\usepackage{amsmath}
\usepackage{bm}
\usepackage{algorithmic}
\usepackage[lined,boxruled,linesnumbered]{algorithm2e}
\usepackage{amssymb}
\usepackage{enumerate}
\usepackage{enumitem}
\usepackage{bbm}
\usepackage{url}
\usepackage{subcaption}
\usepackage[table]{xcolor}
\usepackage{array,hhline}
\usepackage{authblk}
\makeatletter
\newcommand*{\ccol}[1]{%
  \ifdim#1pt<.5pt\relax\else\color{white}\fi
  \edef\x{\noexpand\cellcolor[gray]{\strip@pt\dimexpr1pt-#1pt}}\x
  #1%
}
\newlength{\cellwidth}
\makeatother

\def \rb {\bm{r}} 
\def \cb {\bm{c}} 
\def \hb {\bm{h}} 
\def \wb {\bm{w}} 

\usepackage[pagebackref=true,breaklinks=true,letterpaper=true,colorlinks,bookmarks=false]{hyperref}

 \cvprfinalcopy 

\def\cvprPaperID{442} 

\begin{document}

\title{Joint Inference of Groups, Events and Human Roles in Aerial Videos}


\author{Tianmin Shu$^1$, Dan Xie$^1$, Brandon Rothrock$^2$, Sinisa Todorovic$^3$ and Song-Chun Zhu$^1$\\
$^1$ Center for Vision, Cognition, Learning and Art, University of California, Los Angeles\\
{\tt\small \{stm512, xiedan\}@g.ucla.edu sczhu@stat.ucla.edu}\\
$^2$Jet Propulsion Laboratory, California Institute of Technology\\
{\tt\small brandon.rothrock@jpl.nasa.gov}\\
$^3$School of Electrical Engineering and Computer Science, Oregon State University\\
{\tt\small sinisa@onid.orst.edu}
}

\maketitle

\begin{abstract}
With the advent of drones, aerial video analysis becomes increasingly important; yet, it has received scant attention in the literature. This paper addresses a new problem of parsing low-resolution aerial videos of large spatial areas, in terms of 1) grouping, 2) recognizing events and 3) assigning roles to people engaged in events.  We propose a novel framework aimed at conducting joint inference of the above tasks, as reasoning about each in isolation typically fails in our setting. Given noisy tracklets of people and detections of large objects and scene surfaces (\eg, building, grass), we use a spatiotemporal AND-OR graph to drive our joint inference, using Markov Chain Monte Carlo and dynamic programming. We also introduce a new formalism of spatiotemporal templates characterizing latent sub-events. For evaluation, we have collected and released a new aerial videos dataset using a hex-rotor flying over picnic areas rich with group events. Our results demonstrate that we successfully address above inference tasks under challenging conditions.
\end{abstract}

\begin{figure}
\begin{center}
\includegraphics[width=0.8\linewidth]{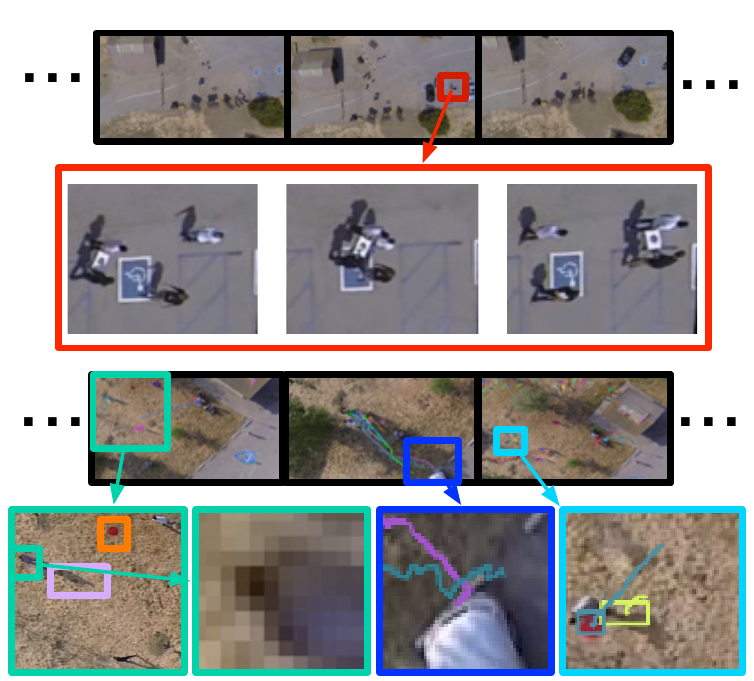}
\end{center}
\caption{Our low-resolution aerial videos show top-down views of people engaged in a number of concurrent events, under camera motion. Different types of challenges are color-coded. The red box marks a zoomed-in video part with varying dynamics among people and their roles \textit{Deliverer} and \textit{Receiver} in \textit{Exchange Box}. The green marks extremely low resolution and shadows. The blue indicates only partially visible \textit{Car}. The cyan marks noisy tracking of person and the small object \textit{Frisbee}.}
\label{fig:fig1}
\end{figure}

\begin{figure*}[t]
\begin{center}
\includegraphics[width=0.95\linewidth]{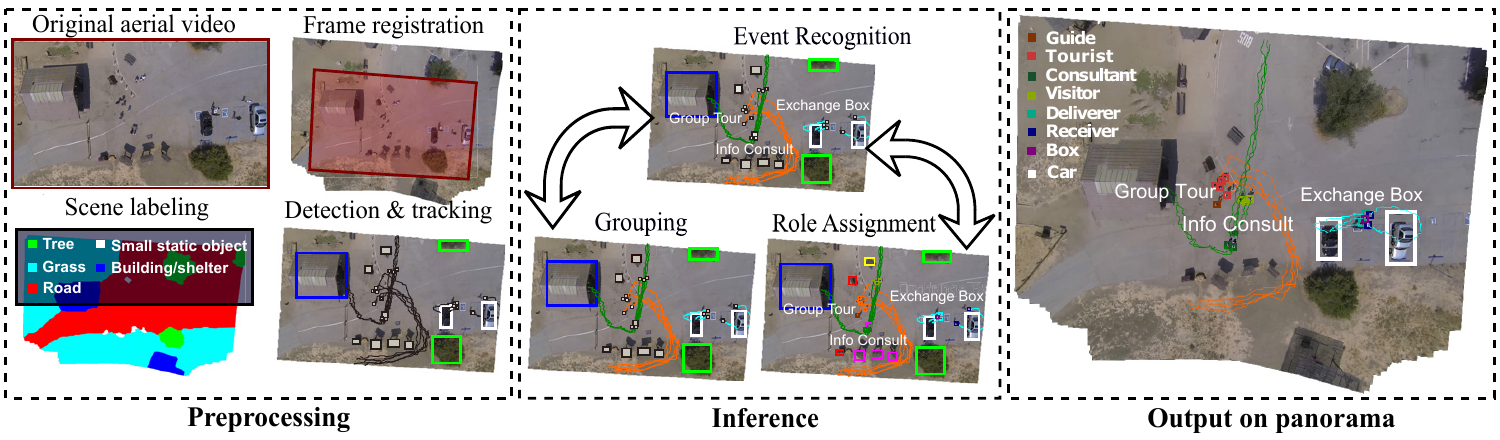}
\end{center}
\caption{The main steps of our approach. Our recognition accounts for the temporal layout of latent sub-events, people's roles within events (\eg, \textit{Guide}, \textit{Visitor}), and small objects that people interact with (\eg, \textit{Box}, \textit{trash bin}). We iteratively optimize groupings of the foreground trajectories, infer their events and human roles (color-coded tracks) within events.}
\label{fig:fig2}
\end{figure*}

\section{Introduction}
\subsection{Motivation and Objective}
Video surveillance of large spatial areas using unmanned aerial vehicles (UAVs) becomes increasingly important in a wide range of civil, military and homeland security applications. For example, identifying suspicious human activities in aerial videos has the potential of saving human lives and preventing catastrophic events. Yet, there is scant prior work on aerial video analysis \cite{Keck2013, Iwashita2013, Prokaj2014}, which for the most part is focused on tracking people and vehicles (with few exceptions \cite{Oreifej2010}) in relatively sanitized settings.

Towards advancing aerial video understanding, this paper presents a new problem of parsing extremely low-resolution aerial videos of large spatial areas, such as picnic areas rich with co-occurring group events, viewed top-down under camera motion, as illustrated in Fig.~\ref{fig:fig1}~and~\ref{fig:fig2}. Given an aerial video, our objectives include:
\begin{enumerate}[itemsep=-3pt,topsep=2pt, partopsep=1pt]
\item Grouping people based on their events;
\item Recognizing events present in each group;
\item Recognizing roles of people involved in these events.
\end{enumerate}

\subsection{Scope and Challenges}\label{sec:challenges}
As illustrated in Fig.~\ref{fig:fig1}, we focus on videos of relatively wide spatial areas (\eg, parks with parking lots) with interesting terrains, taken on-board of a UAV flying at a large altitude (25m) from the ground. People in such videos are formed into groups engaged in different events, involving complex $n$-ary interactions among themselves (\eg, a \textit{Guide} leading \textit{Tourist}s in \textit{Group Tour}), as well as interactions with objects (\eg, \textit{Play Frisbee}). Also, people  play particular roles in each event (\eg, \textit{Deliverer} and \textit{Receiver} roles in \textit{Exchange Box}). 

1. {\bf Low resolution.} People and their portable objects are viewed at an extremely low resolution. Typically, the size of a person is only $15\times 15$ pixels in a frame, and small objects critical for distinguishing one event from another may not be even distinguishable by a human eye. 

2. {\bf Camera motion} makes important cues for event recognition (\eg, object like \textit{Car}) only partially visible or even out of view, and thus may require seeing longer video footage for their reliable detection.

3. {\bf Shadows in top view} make background subtraction very challenging.

Unfortunately, popular appearance-based approaches to detecting people and objects used to produce input for recognizing group events and interactions  \cite{Pellegrini2009, ChoiPAMI2014, Ryoo2011, TianCVPR2012, Ramananthan2013, Fathi2012} do not handle the above three challenges. Thus we have to depart from the appearance-based event recognition. 

In addition, in the face of these challenges, the state of the art methods in people and vehicle tracking frequently miss to track moving foreground, and typically produce short, broken tracklets with a high rate of switched track IDs.

4. {\bf Space-time dynamics.} Our events are characterized by both very large and very small space-time dynamics within a group of people. For example, in the event of a line forming in front of a vending machine, called \textit{Queue for Vending machine}, the participants may be initially scattered across a large spatial area, and may form the line very slowly, while partially occluding one another when closely standing in the line.

\begin{figure*}
\begin{center}
\includegraphics[width=0.95\linewidth]{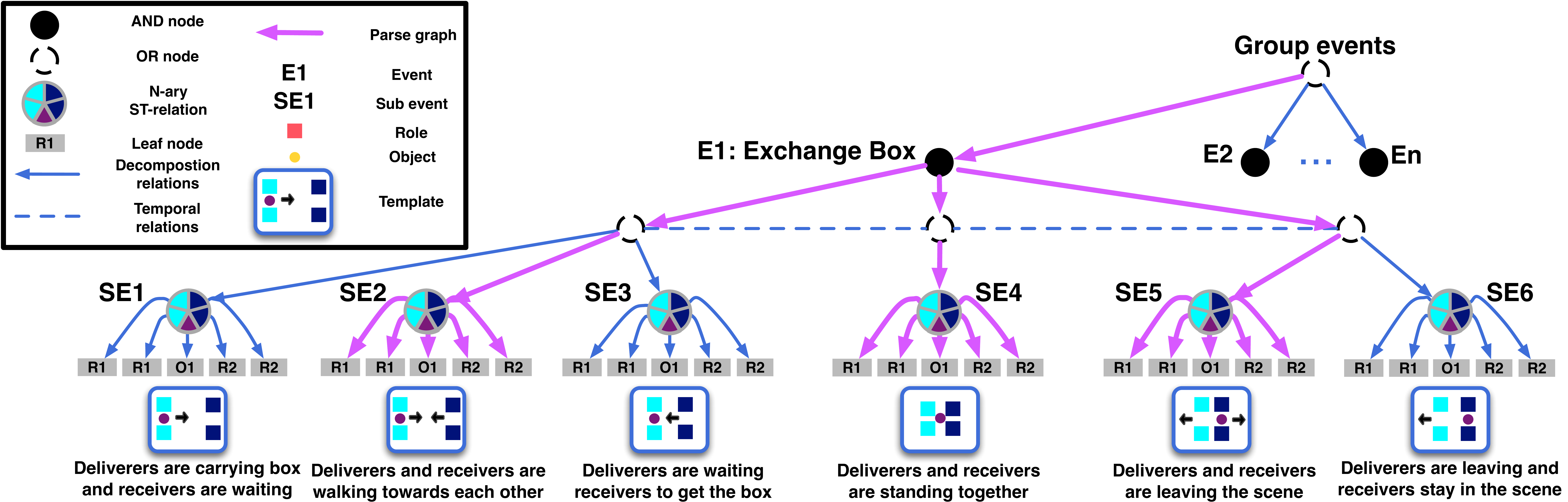}
\end{center}
\caption{A part of ST-AOG for \textit{Exchange Box}. The nodes are hierarchically connected (solid blue) into three levels, where the root level corresponds to events, middle level encodes sub-events, and leaf level is grounded onto foreground tracklets and small static objects in the video. The lateral connections (dashed blue) indicate temporal relations of sub-events. The colored pie-chart nodes represent templates of $n$-ary spatiotemporal relations among human roles and objects (see Fig.~\ref{fig:template}). The magenta edges indicate an inferred parse graph which recognizes and localizes temporal extents of events, sub-events, human roles and objects in the video.}
\label{fig:AOG}
\end{figure*}

\subsection{Overview of Our Approach}
As Fig.~\ref{fig:fig2} illustrates, our approach consists of two main steps:

1. {\bf Preprocessing.} We ground our approach onto noisy detections and tracking. Foreground tracking under camera motion is made feasible by registering video frames onto a reference plane. By frame registration, we generate a panorama for scene labeling. Due to the challenges mentioned in Sec.~\ref{sec:challenges}, tracking of small portable objects and people produces highly unreliable frequently broken tracklets, with a high miss rate. We improve the initial tracking results by agglomeratively clustering tracklets into longer trajectories based on their spatial layout and velocity. We detect large objects (\eg. buildings, cars) using the approach of \cite{Rothrock2013}, and classify superpixels \cite{Achanta2012} of the panorama for scene labeling.

2. {\bf Inference.} We seek event occurrences in the space-time patterns of the foreground trajectories and their relations with the detections of objects in the scene. To constrain our recognition hypotheses under uncertainty, we resort to domain knowledge represented by a probabilistic grammar -- namely, a spatiotemporal AND-OR graph (ST-AOG).  ST-AOG encodes decompositions of events into temporal sequences of sub-events. Sub-events are defined by our new formalism called {\em latent spatiotemporal templates} of $n$-ary relations among people and objects. The templates jointly encode varying spatiotemporal relations of characteristic roles of all people, as well as their interactions with objects, while engaged in the event. 

We specify an iterative algorithm based on Markov Chain Monte Carlo (MCMC \cite{Lee_TPAMI2012}) along with dynamic programming (DP) to jointly infer groups, events and human roles.

\subsection{Prior Work and Our Contributions}\label{sec:related_work}
Our work is related to three research streams.

\textbf{Event Recognition in Aerial Videos}. Prior work on aerial image and video understanding typically puts restrictions on their settings for limited tasks. For example, \cite{Pollard2012} requires robust motion segmentation and learning of object shapes for tracking objects; \cite{Iwashita2013} recognizes people based on background subtraction and motion; and \cite{Prokaj2014} depends on appearance-based regressor and background subtraction for tracking vehicles. Regarding the objectives, these approaches mainly focus on detecting and tracking people or vehicles \cite{Xiao2010, Oreifej2010, Keck2013}. We advance prior work by relaxing their assumptions about the setting, and by extending their objectives to jointly infer groups, events, human roles.

\textbf{Group Activity Recognition}. Simultaneous tracking of multiple people, discovering groups of people, and recognizing their collective activities have been addressed only in every-day videos, rather than aerial videos \cite{Wongun2009, Ryoo2011, Tian2012, Ge2012, Li2013, ChoiPAMI2014, Choi2014, Antic2014, Sun2014, Tu2014}. Also, work on recognizing group activities in large spatial scenes requires high-resolution videos for a ``digital zoom-in'' \cite{Amer2012}. As input, these approaches use person detections along with cues about human appearance, pose, and orientation --- i.e., information that cannot be reliably extracted from our aerial videos. There are also some trajectory-based methods for event recognition \cite{Nevatia2003, Swears2014, loy2012}, but they focus on simpler events compared to what we discuss in this paper. Regarding the representation of collective activities, prior work has used a descriptor of human locations and orientations, similar to shape-context \cite{ChoiPAMI2014, Antic2014}. We advance prior work with our new formalism of latent spatiotemporal template of human roles and their interactions with other actors and objects.

\textbf{Recognition of Human Roles}. Existing work on recognizing social roles and social interactions of people typically requires perfect tracking results \cite{Ramananthan2013}, reliable estimation of face direction and attention in 3D space \cite{Fathi2012}, detection of agent's feet location in the scene \cite{Zhang2011}, and thus are not applicable to our domain. Our approach is related to recent approaches aimed at jointly recognizing events and social roles by identifying interactions of sub-groups  \cite{Ge2012, Li2013, TianCVPR2012, Kwak2013}.

{\bf Contributions:} 
\begin{enumerate}[itemsep=-3pt,topsep=2pt, partopsep=1pt]
\item Addressing a more challenging setting of aerial videos;  
\item New formalism of latent spatiotemporal templates of $n$-ary relations among human roles and objects;
\item Efficient inference using dynamic programming aimed at grouping, recognition and localizing temporal extents of events and human roles 
\item New dataset of aerial videos with per-frame annotations of people's trajectories, object labels, roles, events and groups.
\end{enumerate}


\section{Representation}\label{sec:grammar}
\subsection{Representing of Group Events by ST-AOG}

Similar with hierarchical representation in \cite{Gupta2009, Lin2009, Pei2013, Pirsiavash2014}, domain knowledge is formalized as ST-AOG, depicted in Fig.~\ref{fig:AOG}. Its nodes represent the following four sets of concepts: events $\Delta_\text{E}=\{E_i\}$; sub-events $\Delta_{L}=\{L_a\}$;  human roles $\Delta_\text{R}=\{R_j\}$; small objects that people interact with  $\Delta_\text{O} = \{O_j\}$; and large objects and scene surfaces $\Delta_\text{S}=\{S_j\}$. A particular pattern of foreground trajectories observed in a given time interval gives rise to a sub-event, and a particular sequence of sub-events defines an event.

Edges of the ST-AOG represent decomposition and temporal relations in the domain. In particular,  the nodes are hierarchically connected by decomposition edges into three levels, where the root level corresponds to events, middle level encodes sub-events, and leaf level is grounded onto foreground tracklets and object detections in the video. The nodes of sub-events are also laterally connected for capturing ``followed-by" temporal relations of sub-events within the corresponding events.  


ST-AOG has special types of nodes. An AND node, $\wedge$, encodes a temporal sequence of latent sub-events required to occur in the video so as to enable the event occurrence (\eg, in order to \textit{Exchange Box}, the \textit{Deliverer}s first need to approach the \textit{Receiver}s, give the \textit{Box} to the \textit{Receiver}s, and then leave). For a given event, an OR node, $\vee$, serves to encode alternative space-time patterns of distinct sub-events. 

\subsection{Sub-events as Latent Spatiotemporal Templates}\label{sec:extracting_templates}

\begin{figure}
\begin{center}
\includegraphics[width=0.8\linewidth]{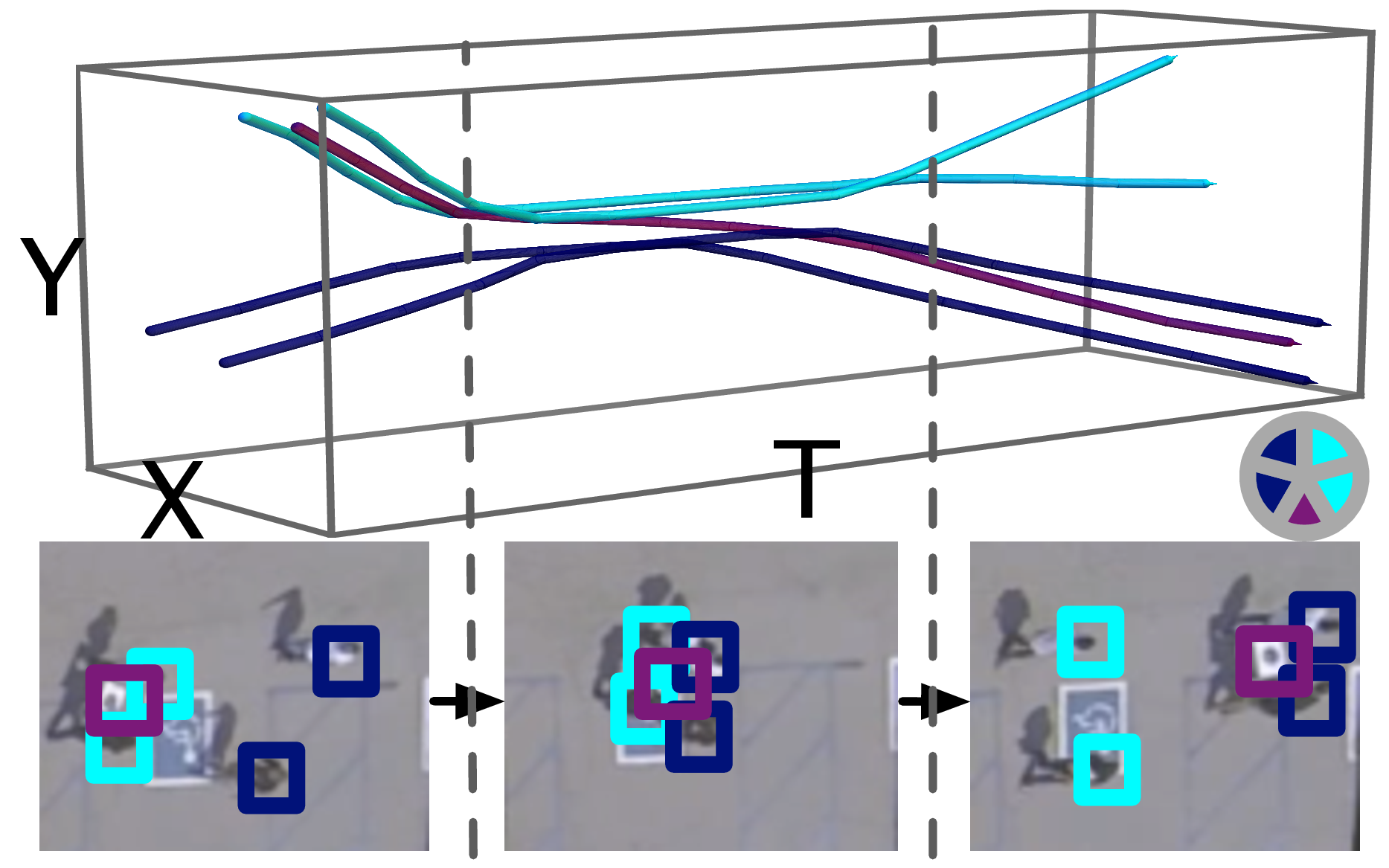}
\end{center}
\caption{Three example templates of $n$-ary spatiotemporal relations among foreground trajectories extracted from the video (XYT-space) for the event \textit{Exchange Box}. The recognized roles \textit{Deliverers}, \textit{Receivers} and the object \textit{Box} in each template are marked cyan, blue and purple, respectively. Spatiotemporal templates are depicted as colored pie-chart nodes in Fig.~\ref{fig:AOG}.}
\label{fig:template}
\end{figure}

A temporal segment of foreground trajectories corresponds to a sub-event. ST-AOG represents a sub-event as the {\em latent} spatiotemporal template of $n$-ary spatiotemporal relations among foreground trajectories within a time interval, as illustrated in Fig.~\ref{fig:template}. In particular, as an event is unfolding in the video, foreground trajectories form characteristic space-time patterns, which may not be semantically meaningful. As they frequently occur in the data, they can be robustly extracted from training videos through unsupervised clustering. Our spatiotemporal templates formalize these patterns within the Bayesian framework using unary, pairwise, and $n$-ary relations among the foreground trajectories. In addition, our unsupervised learning of spatiotemporal templates address unstructured events in a unified manner. Namely, more structured events need more templates and an unstructured one is represented by a single template.

{\bf Unary attributes.} A foreground trajectory,  $\Gamma = [\Gamma^1,...,\Gamma^k,...]$, can be viewed as spanning a number of time intervals, $\tau_k = [t_{k-1},t_{k}]$, where $\Gamma^k = \Gamma(\tau_k)$. Each trajectory segment,  $\Gamma^k$, is associated with unary attributes, $\bm{\phi}=[\rb^k, s^k, \cb^k]$. Elements of the role indicator vector $\rb^k(l)=1$ if $\Gamma^k$ belongs to a person with role $l \in \Delta_\text{R}$ or object class $l\in \Delta_\text{O}$; otherwise $\rb^k(l)=0$. The speed indicator $s^k=1$ when the normalized speed of $\Gamma^k$ is greater than a threshold (we use 2 pixels/sec); otherwise, $s^k = 0$. Elements of the closeness indicator vector $\cb^k(l)=1$ when $\Gamma^k$ is close to any of the large objects or types of surfaces detected in the scene indexed by $l\in \Delta_\text{S}$, such as \textit{Building}, \textit{Car}, for a threshold (70 pixels); o.w., $\cb^k(l) = 0$.

{\bf Pairwise relations.} of a pair of trajectory segments, $\Gamma_j^k$ and $\Gamma_{j'}^k$, are aimed at capturing spatiotemporal relations of human roles or objects represented by the two trajectories, as illustrated in Fig.~\ref{fig:template}. The pairwise relations are specified as: $\bm{\phi}_{jj'}=[d_{jj'}^k, \theta_{jj'}^k, \rb_{jj'}^k, s_{jj'}^k, \cb_{jj'}^k]$, where $d_{jj'}^k$ is the mean distance between $\Gamma_j^k$ and $\Gamma_{j'}^k$; $\theta_{jj'}^k$ is the angle subtended between $\Gamma_j^k$ and $\Gamma_{j'}^k$; and the remaining three pairwise relations check for compatibility between the aforementioned binary relations as: $\rb_{jj'}^k=\rb_j^k\oplus \rb_{j'}^k$, $s_{jj'}^k=s_j^k \oplus s_{j'}^k$, $\cb_{jj'}^k=\cb_j^k \oplus \cb_{j'}^k$, where $\oplus$ denotes the Kronecker product.

 {\bf $n$-ary relations.} Towards encoding unique spatiotemporal patterns of a set of trajectories, we specify the following $n$-ary attribute. A set of trajectory segments, $G_i(\tau_k)=G_i^k=\{\Gamma_j^k\}$, can be described by a 18-bin histogram $\hb^k$ of their velocity vectors. $\hb^k$ counts orientations of velocities at every point along the trajectories in a polar coordinate system: 6 bins span the orientations in $[0,2\pi]$, and 3 bins encode the locations of trajectory points relative to a given center. As the polar-coordinate origin, we use the center location of a given event in the scene.  
 
 {\bf Unsupervised Extraction of Templates.} Given training videos with ground-truth partition of all their ground-truth foreground trajectories $G$ into disjoint subsets $G=\{G_i\}$. Every $G_i$ can be further partitioned into equal-length time intervals  $G_i =\{G_i^k\}$ ($|\tau^k|=2 \text{sec}$). We use K-means clustering to group all  $\{\Gamma_{i,j}^k\}$, and then estimate spatiotemporal templates $\{L_{a}\}$ as representatives of the resulting clusters $a$. For K-means clustering, we use ground-truth values of the aforementioned unary and pairwise relations of $\{\Gamma_{i,j}^k\}$. In our setting of 11 categories of events occurring in aerial videos, we estimate  $|\Delta_{L}|=27$ templates.

\section{Formulation and Learning of Templates}\label{sec:templates}

Given the spatiotemporal templates, $\Delta_{L}=\{L_a\}$, extracted by K-means clustering from training videos (see Sec.~\ref{sec:extracting_templates}), we will conduct inference by seeking these latent templates in foreground trajectories of the new video.
To this end, we define the log-likelihood of a set of foreground trajectories $G=\{\Gamma_{j}\}$ given $L_{a}\in \Delta_{L}$ as
\begin{equation}
\setlength{\arraycolsep}{2pt}
\begin{array}{lcl}
\log p(G|L_{a}) &{\propto}&\displaystyle \sum_j  \wb_{a}^1\cdot\bm{\phi}_j + \sum_{jj'}  \wb_{a}^2\cdot\bm{\phi}_{jj'} + \wb_{a}^3\cdot \hb,\\
&=& \displaystyle \wb_{a}\cdot [\sum_{j} {\phi}_{j}, \sum_{jj'}\bm{\phi}_{jj'},  \hb]= \wb_{a}\cdot \bm{\psi}.
\end{array}
\label{eq:template-posterior}
\end{equation}
where the bottom equation of (\ref{eq:template-posterior}) formalizes every template as a set of parameters $\wb_{a}=[\wb_{a}^1, \wb_{a}^2, \wb_{a}^3]$ appropriately weighting the unary, pairwise and $n$-ary relations of $G$, $\bm{\psi}$. Recall that our spatiotemporal templates are extracted from unit-time segments of foreground trajectories in training. Thus, the log-likelihood in (\ref{eq:template-posterior}) is defined only for sets $G$ consisting of unit-time trajectory segments. 

From (\ref{eq:template-posterior}), the parameters $\wb_{a}$ can be learned by maximizing the log-likelihood of $\{\bm{\psi}_a^k\}$ 
extracted from the corresponding clusters $a$ of training trajectories. 

The log-posterior of assigning template $L_{a}$ to longer temporal segments of trajectories, falling in $\tau=(t',t)$, $t'<t$, is specified as
\begin{equation}
\log p(L_{a}(\tau)| G(\tau))  {\propto}  
\sum_{k=t'}^{t}\log p(G^k|L_{a}) + \log p(L_{a}(\tau)) 
\label{eq:template-posterior-interval}
\end{equation}
where $p(L_{a}(\tau))$ is a log-normal prior that $L_{a}$ can be assigned to a time interval of length $|\tau|$. The hyper-parameters of $p(L_{a}(\tau))$ are estimated using the MLE on training data.

\section{Probabilistic Model}\label{sec:parse_graph}

A parse graph is an instance of ST-AOG, explaining the event, sequence of sub-events, and human role and object label assignment. The solution of our video parsing is a set of parse graphs, $W=\{pg_i\}$, where every $pg_i$  explains a subset of foreground trajectories, $G_i\subset G$,  as 
\begin{equation}
pg_i = \{e_i, \tau_i=[t_{i,0},t_{i,T}], \{L(\tau_{i,u})\}, \{\rb_{i,j}\}\}, 
\label{def_pg}
\end{equation}
where $e_i\in\Delta_\text{E}$ is the recognized event  conducted by $G_i$; $\tau_i=[t_{i,0},t_{i,T}]$ is the temporal extent of $e_i$ in the video starting from frame $t_{i,0}$ and ending at frame $t_{i,T}$; $\{L(\tau_{i,u})\}$ are the templates (i.e., latent sub-events) assigned to non-overlapping, consecutive time intervals $\tau_{i,u} \subset \tau_i$, such that $|\tau_i|=\sum_u |\tau_{i,u}|$;  and $\rb_{i,j}$ is the human role or object class assignment to $j$th trajectory $\Gamma_{i,j}$ of $G_i$. 


Our objective is to infer $W$ that maximizes the log-posterior  $\log p(W | G) \propto - \mathcal{E}(W | G)$, given all foreground trajectories $G$ extracted from the video. The corresponding energy $\mathcal{E}(W | G)$ is specified for a given partitioning of $G$ into $N$ disjoint subsets $G_i$ as
\begin{equation}
\setlength{\arraycolsep}{0pt}
\begin{array}{lcl}
\mathcal{E}(W | G)  &{\propto} & \displaystyle
\sum_{i=1}^{N} \Big[-\underbrace{\log p(\wedge_{e_i}|\vee_{\text{root}})}_{\text{select event}~e_i} + \sum_u \big[-\underbrace{\log p(\wedge_{L_a}|\vee_{e_i})}_{\text{select template}~L_a} \\
&&\displaystyle  - \underbrace{\log p(L_a(\tau_{i,u})| G_i(\tau_{i,u}))}_{\text{assign template}}\big]\Big]
\end{array}
\label{eq:energy_W}
\end{equation}
where $G_i(\tau_{i,u})$ denotes temporal segments of foreground trajectories falling in time intervals $\tau_{i,u}$, $|\tau_i|=\sum_{u}|\tau_{i,u}|$, and $\log p(L(\tau_{i,u})| G_i(\tau_{i,u}))$ is given by (\ref{eq:template-posterior-interval}). Also, $\log p(\wedge_{e_i}|\vee_{\text{root}})$ and $\log p(\wedge_{L_a}|\vee_{e_i})$ are the log-probabilities of the corresponding switching OR nodes in ST-AOG for selecting particular events $e_i\in\Delta_E$ and spatiotemporal templates $L_a\in\Delta_L$. These two switching probabilities are simply estimated as the frequency of corresponding selections observed in training data.

\section{Inference}\label{sec:inference}

Given an aerial video, we first build a video panorama and extract foreground trajectories $G$. Then, the goal of inference is to: (1) partition $G$ into disjoint groups of trajectories $\{G_i\}$ and assign label event $e_i\in\Delta_\text{E}$ to every $G_i$; (2) assign human roles and object labels $\rb_{i,j}$ to trajectories $\Gamma_{i,j}$ within each group $G_i$; and 3) assign latent spatiotemporal templates $L(\tau_{i,u})\in \Delta_{L}$ to temporal segments $\tau_{i,u}$ of foreground trajectories within every $G_i$. For steps (1) and (2) we use two distinct MCMC processes. Given groups $G_i$, event labels $e_i$ and role assignment $r_{i,j}$ proposed in (1) and (2), step (3) uses dynamic programming for efficient estimation of sub-events $L(\tau)$ and their temporal extents $\tau$. Steps (1)--(3) are iterated until convergence, i.e., when $\mathcal{E}(W | G)$, given by (\ref{eq:energy_W}), stops decreasing after a sufficiently large number of iterations.

\subsection{Grouping}

Given $G$, we first use \cite{Ge2012} to perform initial clustering of foreground trajectories into atomic groups. Then, we apply the first MCMC to iteratively propose either to merge two smaller groups into a merger, with probability  $p(1)=0.7$,  or to split a merger into two smaller groups, with probability  $p(2)=0.3$. Given the proposal, each resulting group $G_i$ is labeled with an event $e_i\in\Delta_\text{E}$ (we enumerate all possible labels). In each proposal, the MCMC jumps from current solution $W$ to a new solution {$W^{\prime}$} generated by one of the dynamics. The acceptance rate is {$\alpha = \min\left\{1,\frac{Q(W \rightarrow W^{\prime})p(W^{\prime} | G)}{Q\left(W^{\prime} \rightarrow W\right)p\left(W | G\right)}\right\}$}, where the proposal distribution $Q(W \rightarrow W^{\prime})$ is one of  $p(1)$ or $p(2)$ depending on the proposal, and $p\left(W | G\right)$ is given by (\ref{eq:energy_W}). 

\subsection{Human Role Assignment}

Given a partitioning of $G$ into groups $\{G_i\}$ and their event labels $\{e_i\}$, we use the second MCMC process within every $G_i$ to assign human roles and object labels to trajectories. Each trajectory $\Gamma_{i, j}$ in $G_i$ is randomly assigned with an initial human-role/object label $\rb_{i,j}$ for solution $pg_i$. In each iteration, we randomly select $\Gamma_{i, j}$ and change it's role label to generate a new proposal $pg^{\prime}_i$. The acceptance rate is $\alpha = \min\left\{1,\frac{Q(pg_i \rightarrow pg^{\prime}_i)p(pg^{\prime}_i | G_i)}{Q(pg^{\prime}_i \rightarrow pg_i)p\left(pg_i | G_i\right)}\right\}$, where $\frac{Q(pg_i \rightarrow pg^{\prime}_i)}{Q(pg^{\prime}_i \rightarrow pg_i)} = 1$ and $p\left(pg^{\prime}_i | G_i\right)$ is maximized by dynamic programming specified in the next section~\ref{sec:DP}.

\begin{figure}[t]
\begin{center}
 \includegraphics[width=0.95\linewidth]{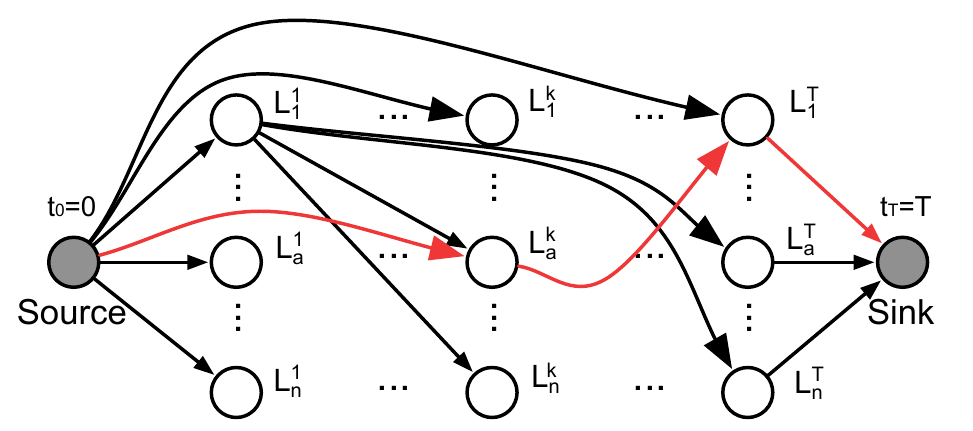}
\end{center}
  \caption{Our DP process can be illustrated by this DAG (directed acyclic graph). An edge between $L_{a'}^{k'}$ and $L_{a}^k$ means the transition $L_{a'} \rightarrow L_{a}$ follows the rule defined in ST-AOG and the time interval $[t_{a'}, t_{a}]$ is assigned with template $L_{a}$. In this sense, with the transition rules and the prior defined in (\ref{eq:template-posterior-interval}) (we do not consider the assignment with low prior probability), we can define the edges of such DAG. So the goal of DP is equivalent to finding a shortest path between source and sink. The red edges highlight a possible path. Suppose we find a path $source \rightarrow L_3^8 \rightarrow L_1^{20} \rightarrow sink$. This means that we decompose $[0, T]$ into 2 time intervals: $[0, 8\delta t], [8 \delta t, T]$, and they are assigned with template $L_3$ and $L_1$ respectively.}
\label{fig:DP}
\end{figure}

\subsection{Detection of Latent Sub-events with DP}
\label{sec:DP}
From steps (1) and (2), we have obtained the trajectory groups $\{G_i\}$, and their event $\{e_i\}$ and role labels $\{\rb_{i,j}\}$. Every $G_i$ can be viewed as occupying time interval of $\tau_i=[t_{i,0}, t_{i,T}]$. The results of steps (1) and (2) are jointly used with detections of large objects $\{S_i\}$ to estimate all unary, pairwise, and $n$-ary relations $\bm{\psi}_i$ of every $G_i$. Then, we apply
dynamic programming for every $G_i$ in order to find latent templates $L(\tau_{i,u})\in\Delta_{L}$ and their optimal durations $\tau_{i,u}\subset [t_{i,0}, t_{i,T}]$. In the sequel, we drop notion $i$ for the group, for simplicity.

The optimal assignment of sub-events can be formulated using a graph, shown in Fig.~\ref{fig:DP}. To this end, we partition $[t_0, t_T]$ into equal-length time intervals $\{[t_{k-1}, t_k]\}$, where $ t_k-t_{k-1} = \delta t$, $\delta t = 2 \text{sec}$. Nodes $L_a^k$ in the graph represent the assignment of templates $L_a\in\Delta_L$ to the intervals $ [t_{k-1},t_k]$. The graph also has the source and sink nodes. 

Directed edges in the graph are established only between nodes $L_a^{k'}$ and $L_a^k$, $1 \leq k' < k$, to denote a possible assignment of  the very same template $L_a$ to the temporal sequence $[t_{k'},t_k]$. The directed edges are assigned weights (a.k.a. belief messages), $m(L_a^{k'}, L_a^k)$, defined as
\begin{equation}
m(L_a^{k'}, L_a^k) = \log p(L_a(t_{k'}, t_{k})| G_i(t_{k'},t_{k})),
\label{eq:edge_weight}
\end{equation}
where  $\log p(L_a(t_{k'}, t_{k})| G_i(t_{k'},t_{k}))$ is given by (\ref{eq:template-posterior-interval}). Consequently, the belief of node $L_a^{k}$ is defined as 
\begin{equation}
\begin{array}{lr}
b(L_a^{k}) = \displaystyle\max_{k', a'} ~ b(L_{a'}^{k'}) + m(L_a^{k'}, L_a^k). & \text{[Forward pass]}
\end{array}
\label{eq:Forward}
\end{equation}

 Here $b(L_a^{0}) = 0$. We compute the optimal assignment of latent sub-events using the above graph in two passes. In the {\em forward pass}, we compute the beliefs of all nodes in the graph using (\ref{eq:Forward}). Then, in the {\em backward pass}, we backtrace the optimal path between the sink and source nodes, in the following steps:
\begin{enumerate}[itemsep=-3pt,topsep=2pt, partopsep=1pt]
\item[0:] Let $t_k\leftarrow t_T$;  
\item[1:] Find the optimal sub-event assignment at time $t_k$ as $L_{a^*}^{k} = \arg\max_{a} ~ b(L_a^k)$; let $a \leftarrow a^*$;
\item[2:] Find the best time moment in the past $t_{k^*}$, $k^*{<}k$, and its best sub-event assignment as $L_{a^*}^{k^*} = \max_{a', k'} b(L_{a'}^{k'}) {+} m(L_a^{k'}, L_a^k)$; Let $a {\leftarrow} a^*$ and $k {\leftarrow} k^*$.
\item[3:]  If $t_k > t_0$, go to Step 2. 
\end{enumerate}

\begin{table*}\footnotesize
\begin{center}
\begin{tabular}{c|c|c|c|c|c} \hline
  & Method  & Input setting & Group & Event & Role \\ \hline
Baseline Var& \cite{Ge2012} for grouping, \cite{ChoiPAMI2014} for event and role classification. & Ground-truth tracks + object annotation & $77.71\%$ & $17.22\%$ & $13.98\%$  \\ \hline
Baseline & Baseline method as above. & Tracking result & $39.64\%$ & $16.94\%$ & $5.53\%$ \\ \hline
Ours Var1 & Our full model & Ground-truth tracks + object annotation & $95.48\%$ & $96.38\%$ & $89.94\%$ \\ \hline
Ours Var2 & Our full model & Tracking result + object annotation & $87.55\%$ & $54.75\%$ & $28.86\%$ \\ \hline
Ours Var3 & Our full model & Tracking result + group labeling & N/A & $39.92\%$ & $18.71\%$ \\ \hline
Ours Var4 & Our model without temporal event grammar & Tracking result & $40.41\%$ & $18.51\%$ & $8.69\%$  \\ \hline
Ours & Our full model & Tracking result  & $49.47\%$ & $32.84\%$ & $18.92\%$ \\ \hline
\end{tabular}
\end{center}
\caption{Comparison of our method with baseline methods and variants of our approach. Our method yields best accuracy based on ground-truth bounding boxes and object labels compared to the baseline methods. Using noisy tracking and object detection results, the accuracy is limited, yet better than the baseline methods under the same condition. This demonstrates the advantages of our joint inference. When given access to the ground-truth of objects or people grouping, our results improve. Without reasoning about latent sub-events, accuracy drops significantly, which justifies our model's ability to capture the structural variations of group events.}
\label{tab:baseline_list}
\end{table*}

\section{Experiment}\label{sec:experiments}

{\bf Existing Datasets.} Existing datasets on aerial videos, group events or human roles are inappropriate for our evaluation. These aerial videos or images indeed show some group events, but the events are not annotated (\cite{Ali2007, AFRL2009, Oreifej2010, Oh2011}). Most aerial datasets are compiled for tracking evaluation only \cite{Keck2013, Iwashita2013, Prokaj2014}. Existing group-activity videos \cite{Wongun2009, Ryoo2011, Amer2012, Li2013} or social role videos \cite{Zhang2011, Fathi2012, TianCVPR2012, Ramananthan2013, Kwak2013}  are captured on or near the ground surface, and have sufficiently high resolution for robust people detection. Thus, we have prepared and released a new aerial video dataset \footnote{Dataset can be download from \url{http://www.stat.ucla.edu/~tianmin.shu/AerialVideo/AerialVideo.html}} with the new challenges listed in Sec.~\ref{sec:challenges}.

{\bf Aerial Events Dataset.} A hex-rotor with a GoPro camera was used to shoot aerial videos at altitude of 25 meters from the ground. The videos show two different scenes, viewed top-down from the flying hex-rotor. The dataset contains 27 videos, 86 minutes, 60 fps, resolution of $1920 \times 1080$, with about 15 actors in each video.  All video frames are registered onto a reference plane of the video panorama.  Annotations are provided (\cite{Vondrick2012}) as: bounding boxes around groupings of people, events, human roles, and small and large objects. The objects include: 1. \textit{Building}, 2. \textit{Vending Machine}, 3. \textit{Table \& Seat}, 4. \textit{BBQ Oven}, 5. \textit{Trash Bin}, 6. \textit{Shelter}, 7. \textit{Info Booth}, 8. \textit{Box}, 9. \textit{Frisbee}, 10. \textit{Car}, 11. \textit{Desk}, 12. \textit{Blanket}. The  events include: 1. \textit{Play Frisbee}, 2. \textit{Serve Table}, 3. \textit{Sell BBQ}, 4. \textit{Info Consult}, 5. \textit{Exchange Box}, 6.\textit{ Pick Up}, 7. \textit{ Queue for Vending Machine}, 8. \textit{Group Tour}, 9. \textit{Throw Trash}, 10. \textit{Sit on Table}, 11. \textit{Picnic}. The human roles include: 1. \textit{Player}, 2. \textit{Waiter}, 3. \textit{Customer}, 4. \textit{Chef}, 5. \textit{Buyer}, 6. \textit{Consultant}, 7. \textit{Visitor}, 8. \textit{Deliverer}, 9. \textit{Receiver}, 10. \textit{Driver}, 11. \textit{Queuing Person}, 13. \textit{Guide}, 14. \textit{Tourist}, 15. \textit{Trash Thrower}, 16. \textit{Picnic Person}.

\begin{figure*}
\begin{center}
\includegraphics[width=1.0\linewidth]{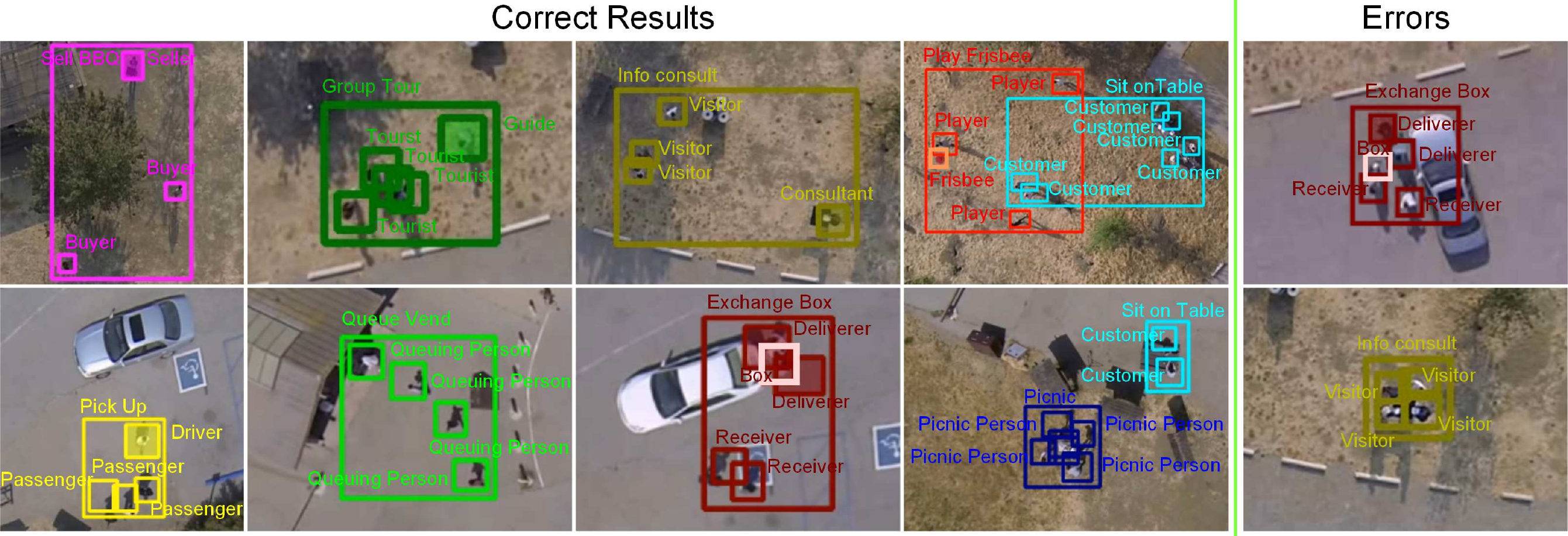}
\end{center}
\caption{Visualization of results including groups (large bounding boxes), events (text) and human roles (small bounding boxes with text). In events with more than one role, we use the shaded bounding box to represent the second role; small portable objects are labeled with lighter color. From event and human role recognition, we can group people even when they are far from each other (\eg,\textit{Play Frisbee} and \textit{Sell BBQ}). In the top-rightmost failure example, true event \textit{Pick Up} is wrongly recognized as \textit{Exchange Box} because one person's trajectory is inferred as \textit{Box}. In bottom-rightmost failure example, our event recognition is correct, but true \textit{Consultant} role is wrongly inferred as \textit{Visitor} role. }
\label{fig:res_datasets}
\end{figure*}

{\bf Evaluation Metrics.} We split the 27 videos into 3 sets, such that different event categories are evenly distributed, and use a three-fold cross validation for our evaluation. Although our training and test videos show the same two scenes, we make the assumption that the layout of ground surfaces and large objects is unknown. Also, different videos in our dataset cover different parts of these large scenes, which are also assumed unknown. We evaluate accuracy of: i) grouping people, ii) event recognition, iii) role assignment. While our approach also estimates sub-events, note that they are latent and not annotated. The results are all time-averaged with the lengths of trajectories in each video. For specifying evaluation metrics we use the following notation. {$G = \{G_i\}$} and {$G^{\prime}=\{G_i^{\prime}\}$} are the sets of groups in ground-truth and inference results respectively. {$\Gamma_{ij}$} is the $j$th trajectory in $i$th group in ground-truth data, with duration of $|\tau_{ij}|$, group label $g_{ij}$, event type $e_{ij}$ and human role $r_{ij}$ in ground-truth. So is {$\Gamma^{\prime}_{ij}$} in our inference. For group $G_i$, we call the best matched (i.e. overlapped) group in $G^{\prime}$ as $M_i$. For group $G^{\prime}_i$, we call the best match group in $G$ as $M^{\prime}_i$. Then, precision and recall of grouping  are
\begin{equation}\footnotesize
Pr_g = \sum\limits_{G_i \in G} \Big({{\sum\limits_{\Gamma_{ij} \in G_i}{\mathbbm{1}\left(M_i = g^{\prime}_{ij}\right)}\cdot |\tau_{ij}|}/{\sum\limits_{\Gamma_{ij} \in G_i}{|\tau_{ij}|}}} \Big)
\end{equation}
\begin{equation}\footnotesize\textstyle
Rc_g = \sum\limits_{G^{\prime}_i \in G^{\prime}} \Big({{\sum\limits_{\Gamma^{\prime}_{ij} \in G^{\prime}_i}{\mathbbm{1}\left(M^{\prime}_i = g_{ij}\right)}\cdot |\tau^{\prime}_{ij}|}/{\sum\limits_{\Gamma^{\prime}_{ij} \in G^{\prime}_i}{|\tau^{\prime}_{ij}|}}} \Big)
\end{equation}
Accuracy of grouping is $F_g = 2 \Big / (1 / Pr_g + 1 / Rc_g)$.

Event recognition accuracy {$E_e$} and role assignment accuracy {$E_r$} are defined as
\begin{equation}\footnotesize
E_e = {\sum\limits_{G_i^{\prime} \in G^{\prime}} \Big( {\sum\limits_{\Gamma^{\prime}_{ij} \in G_i^{\prime}}{\mathbbm{1}\left({e_{ij} = e^{\prime}_{ij}}\right)\cdot |\tau_{ij}|}}\Big)}  /{\sum\limits_{G_{i}^{\prime} \in G^{\prime}}{\sum\limits_{\Gamma^{\prime}_{ij} \in G_i^{\prime}}{|\tau_{ij}|}}}
\end{equation}
\begin{equation}\footnotesize
E_r = {\sum\limits_{G_i^{\prime} \in G^{\prime}} \Big( \sum\limits_{\Gamma^{\prime}_{ij} \in G_i^{\prime}}{\mathbbm{1}\left({r_{ij} = r^{\prime}_{ij}}\right)\cdot |\tau_{ij}|}\Big)}  /{\sum\limits_{G_{i}^{\prime} \in G^{\prime}}{\sum\limits_{\Gamma^{\prime}_{ij} \in G_i^{\prime}}{|\tau_{ij}|}}}.
\end{equation}\par


\begin{figure*}
\begin{center}
\begin{subfigure}[b]{0.33\linewidth}
\includegraphics[width=1\linewidth]{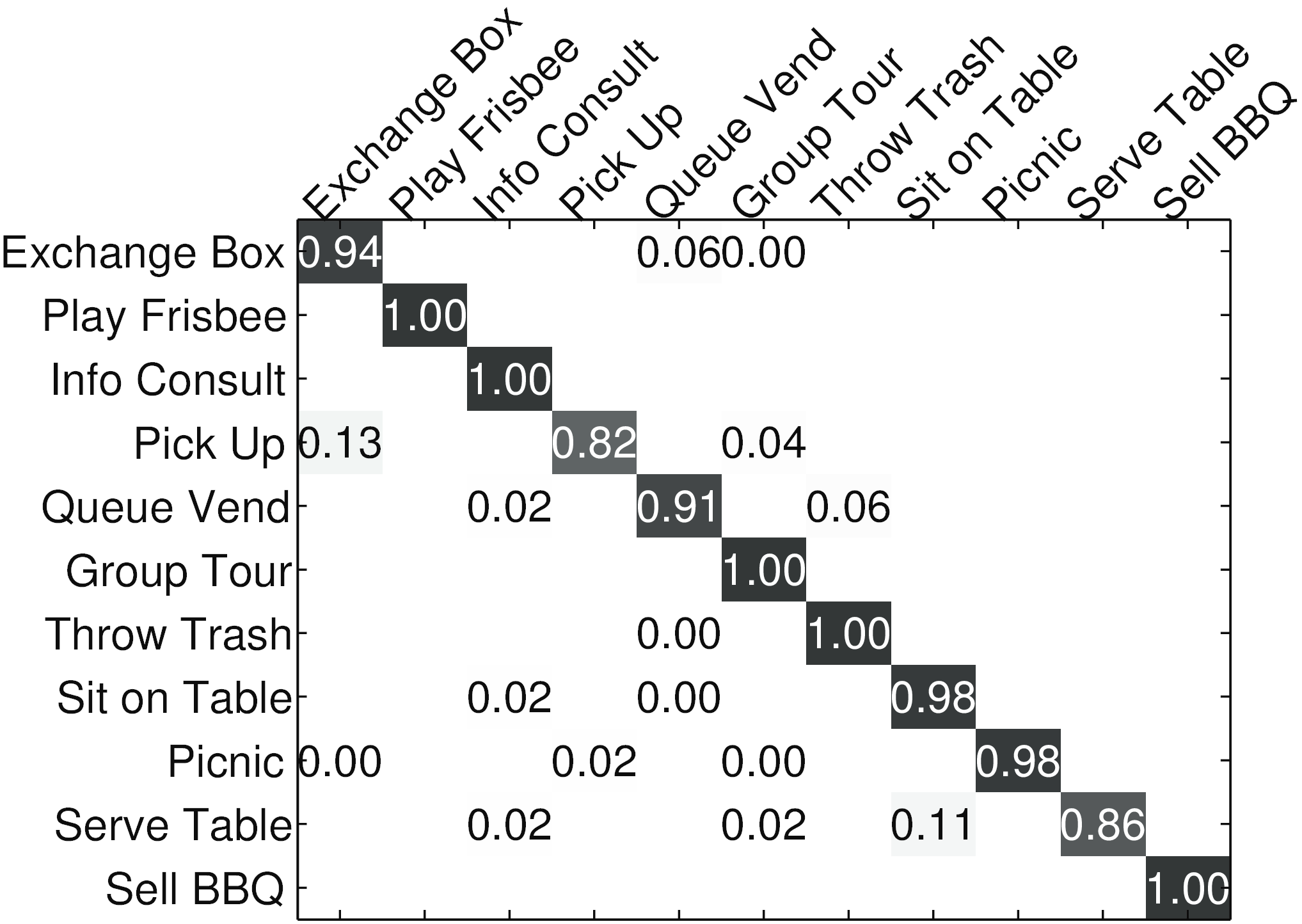}
\caption{event recognition on GT}
\end{subfigure}
\begin{subfigure}[b]{0.33\linewidth}
\includegraphics[width=1\linewidth]{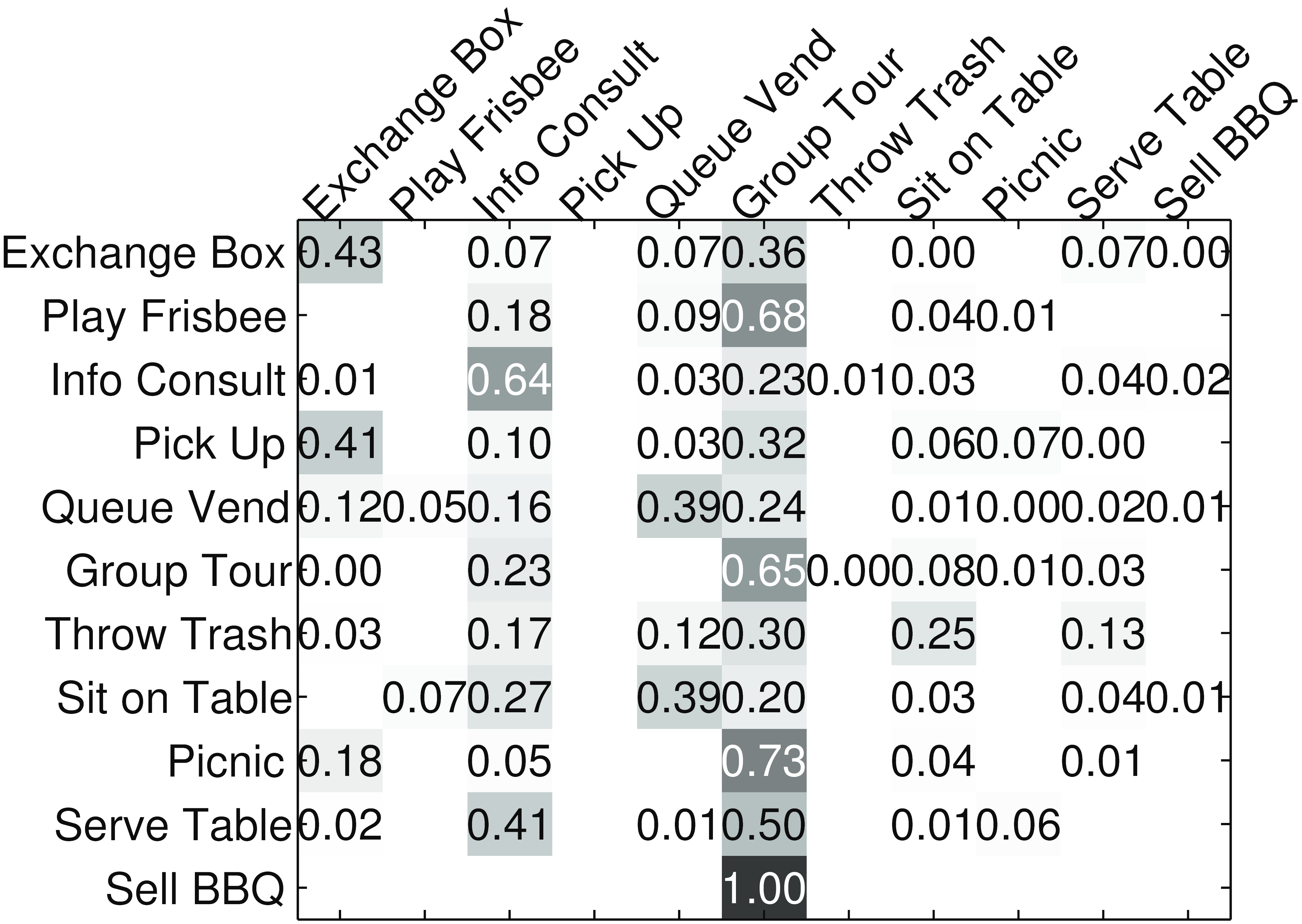}
\caption{event recognition on tracking result}
\end{subfigure}
\begin{subfigure}[b]{0.33\linewidth}
\includegraphics[width=1\linewidth]{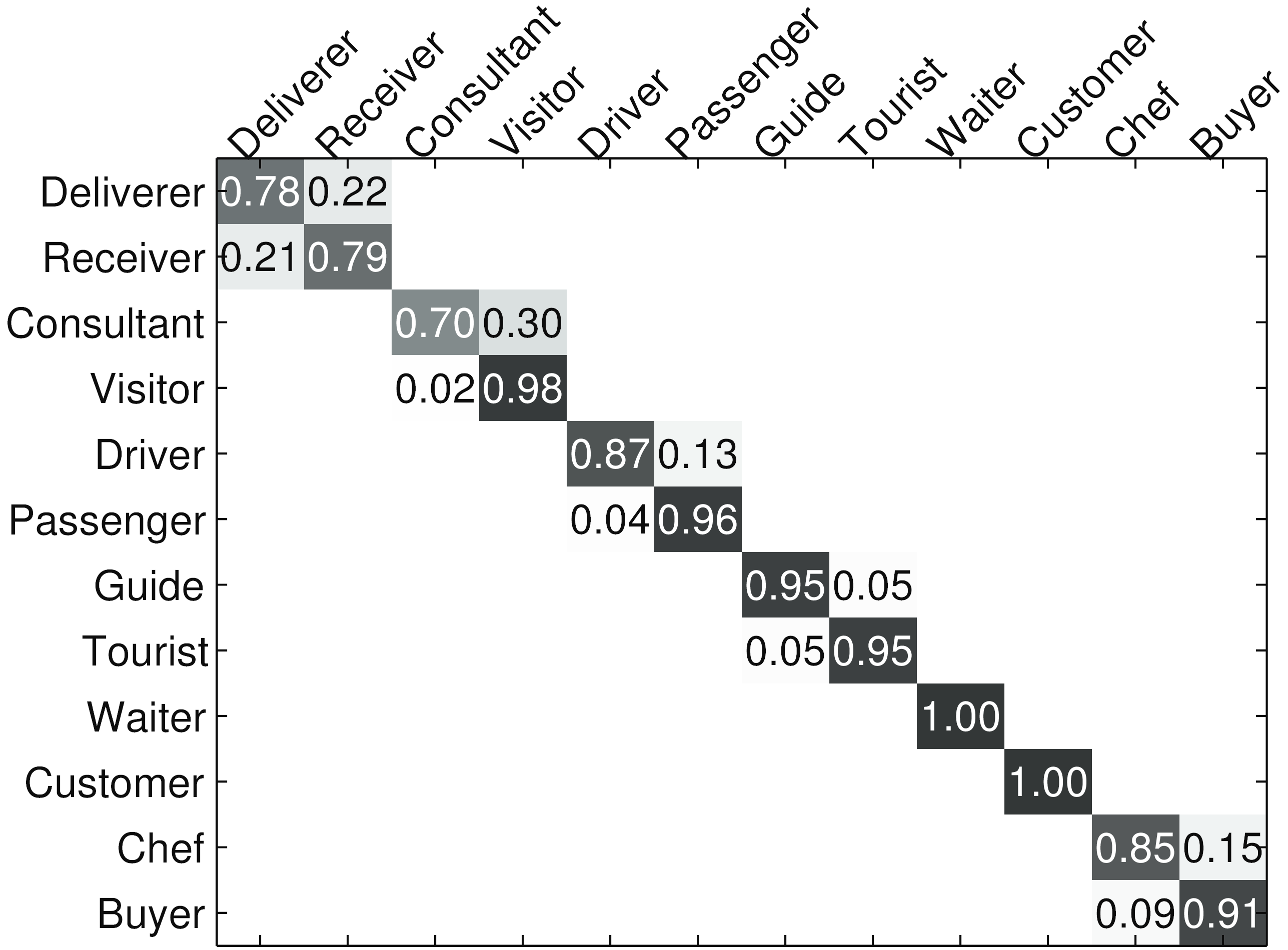}
\caption{role assignment on GT}
\end{subfigure}
\end{center}
\caption{Confusion matrices of event recognition and role assignment result. (a) is event recognition result based on ground-truth (GT) bounding boxes and object labels; (b) is result based on real tracking and detections. From (a) and (b) we can see that \textit{Info Consult}, \textit{Sit on Table}, \textit{Serve Table} cannot be easily distinguished from each other solely based on noisy tracklets. Some events (\eg \textit{Group Tour}) tend to be wrongly favored by our approach, especially when we do not observe some distinguishing objects. (c) is role assignment result confusion matrix within event class  based on ground-truth bounding boxes and object labels. Each $2 \times 2$ block is a confusion matrix of role assignment within that event.}
\label{fig:confusion_mat_event}
\end{figure*}


{\bf Baselines.} To evaluate effectiveness of each module of our approach, we compare with baselines and variants of our method defined in Tab.~\ref{tab:baseline_list}. For the baselines we extract the following low-level features on trajectories: shape-context like feature \cite{Wongun2009}, average velocity, aligned orientation, distance from each type of large objects. All elements of feature vectors are normalized to fall in [0, 1].

{\bf Results}. We register raw videos by RANSAC over Harris Corner feature points, then apply method of \cite{Iwashita2013} for tracking, which is based on background subtraction \cite{Yao2007, bgslibrary}. We also use the detector of \cite{Rothrock2013} to detect buildings and cars, while other static objects are inferred in scene labeling. We do not detect portable objects, \eg, \textit{Frisbee} and \textit{Box}.

We evaluate our approach on both annotated bounding boxes and real tracking results. Example qualitative results are presented in Fig. \ref{fig:res_datasets}. As can be seen, the results are reasonably good. The quantitative results are shown in Tab. \ref{tab:baseline_list}. Confusion matrices of event recognition and role assignment are shown in Fig. \ref{fig:confusion_mat_event}. Additional results are presented in the supplementary material.


\section{Conclusion}
We collected a new aerial video dataset with detailed annotations, which presents new challenges to computer vision and complements existing benchmarks. We specified a framework for joint inference of events, human roles and people groupings using noisy input. Our experiments showed that addressing each of these inference tasks in isolation is very difficult in aerial videos, and thus provided justification for our holistic framework. Our results demonstrated significant performance improvements over baselines when we constrained uncertainty in input features with domain knowledge.

Our model is limited and can be extended in two directions. First, we infer the function of the objects implicitly based on the group events currently. In the future, we wish to explicitly infer the functional map for a given site, in the sense that certain area corresponds to specific human activities, e.g., dinning area, parking lot, etc. Unlike appearance-based aerial image parsing \cite{Porway2010}, the spatial segmentation will be guided by the spatiotemporal characteristics of human activities. Second, similar to what \cite{Xie2013} did for the prediction of individual intention, we would like to reason the intention of a group as another extension of our work.


\section*{Acknowledgements}
This research has been sponsored in part by grants DARPA MSEE FA 8650-11-1-7149, ONR MURI N00014-10-1-0933 and NSF IIS-1423305. The authors would like to thank Dr. Michael Ryoo at JPL for the helpful discussions.


\bibliographystyle{ieee}\small
\bibliography{AerialVideo}

\end{document}